\documentclass[conference]{IEEEtran}
\ifCLASSINFOpdf
   \usepackage[pdftex]{graphicx}

\else
\fi
\usepackage[cmex10]{amsmath}

\DeclareMathOperator*{\argmax}{arg\,max}
\renewcommand{\vec}[1]{\mathbf{#1}}

\begin{document}
\title{A pragmatic approach to multi-class classification}

\author{\IEEEauthorblockN{Thomas Kopinski, St\'ephane Magand, Uwe Handmann}
\IEEEauthorblockA{Hochschule Ruhr-West\\
L\"utzowstra\ss e 5\\
46236 Bottrop, Germany\\
Email: firstname.lastname@hochschule-ruhrwest.de}
%\and
%\IEEEauthorblockN{St\'ephane Magand}
%\IEEEauthorblockA{Hochschule Ruhr-West\\
%L\"utzowstraße 5\\
%46236 Bottrop, Germany\\
%Email: stephane.magand@\\hochschule-ruhrwest.de}
\and
\IEEEauthorblockN{Alexander Gepperth}
\IEEEauthorblockA{ENSTA ParisTech, INRIA FLOWERS\\
828 Blvd des Mar\'echaux\\
91762 Palaiseau, France\\
Email: alexander.gepperth@ensta.fr}}


%Telephone: (800) 555--1212, Fax: (888) 555--1212}


\maketitle

\begin{abstract}
We present a novel hierarchical approach to multi-class classification which is generic in that it can be applied to different classification models (e.g., support vector machines, perceptrons), and makes no explicit assumptions about the probabilistic structure of the problem as it is usually done in multi-class classification. By adding a cascade of additional classifiers, each of which receives the previous classifier's output in addition to regular input data, the approach harnesses unused information that manifests itself in the form of, e.g., correlations between predicted classes. Using multilayer perceptrons as a classification model, we demonstrate the validity of this approach by testing it on a complex ten-class 3D gesture recognition task.
\end{abstract}


% \fi
\IEEEpeerreviewmaketitle

\section{Introduction}
% Context
This contribution is in the context of classification, in particular multi-class classification (MCC) where an input data vector has to be assigned one 
of multiple output classes. Although there is a truly abundant amount of work on binary classification, its foundations and applications, 
the same cannot be said for MCC. This is probably due to the fact that the statistical theory behind the binary case\cite{Bishop2006} cannot be 
trivially generalized\cite{guermeur}, making model and hyperparameter selection much more complex to treat in MCC. On a purely application-oriented level, there is a multitude of models that have been proposed for performing MCC, although reported performances are in general very similar, or, where they differ, the differences are strongly task-dependent and show no trend towards universally "better" or "worse" methods. Additionally, almost all of the proposed methods suffer either from
prohibitive training complexity, unclear assumptions on the problem, or difficult-to-tune parameters. 
\subsection{Related work}
When it comes to MCC, generally there is a distinction between models that can directly treat multiple classes, and decomposition approaches that reduce a multi-class problem to several binary ones. "Direct" models include extensions of large margin classifiers such as M-SVMs \cite{msvm-1,msvm-2,msvm-3,msvm-4}, multinomial kernel regression\cite{zhu2005kernel} and even multilayer perceptrons (MLPs) if classification is treated as a regression problem. In this contribution, we focus on decomposition approaches as there seems to be no evidence at all that they perform worse than "direct" MCC\cite{guermeur}, and in fact are often much more computationally efficient\cite{guermeur}. There has been a considerable body of work on decomposition approaches for support vector machines (SVMs) which are very powerful binary classifiers. Nevertheless, since such decomposition approaches are in principle independent of the choice of the underlying classifier, as long as it is binary, all results are generalizable. There are two main decomposition approaches: first of all, there is the "one-versus-all" (OVA) \cite{ova-1,ova-2} approach which trains one binary classifier to distinguish one class from all the other classes. Then, the final decision is simply the argmax over all classifier responses. On the other hand, there is the "one-vs-one" (OVO) approach \cite{ovo-1,ovo-2,ovo-3,ovo-4,ovo-5,ovo-6,ovo-7} which trains a binary classifier for each pair of classes. The final decision is then, in the simplest formulation, obtained by a voting scheme among all pairwise classifiers. The main drawback of this approach are the limited number of samples available for each pairwise classifier, (although there are more complex formulations that fix this\cite{ovo-8}), and the assumption that
a pairwise classifier will have a weak response when presented with classes unknown to it. Lastly, there are more complex graph-based approaches\cite{trees-1,trees-2} that construct decision trees, at each node of which there is a binary classifier that determines the 
progress in the tree until a leaf is reached. 

Common to all of the decomposition approaches is the fact that more or less strong assumptions are made about the
problem at hand. Most fundamentally, all approaches need to make responses from different binary classifiers comparable.
As this is not possible in general for discriminatively trained classifiers, some sort of calibration procedure (often the technique from \cite{platt1999probabilistic}) is used to obtain normalized "probabilities" from classifier outputs, although this makes more or less strong assumptions about the data. Secondly, especially for OVO approaches, the precise way of implementing the voting scheme 
makes assumptions (usually about conditional independence properties) that may or may not apply. 
\begin{figure}
\centering
\includegraphics[width=0.3\textwidth]{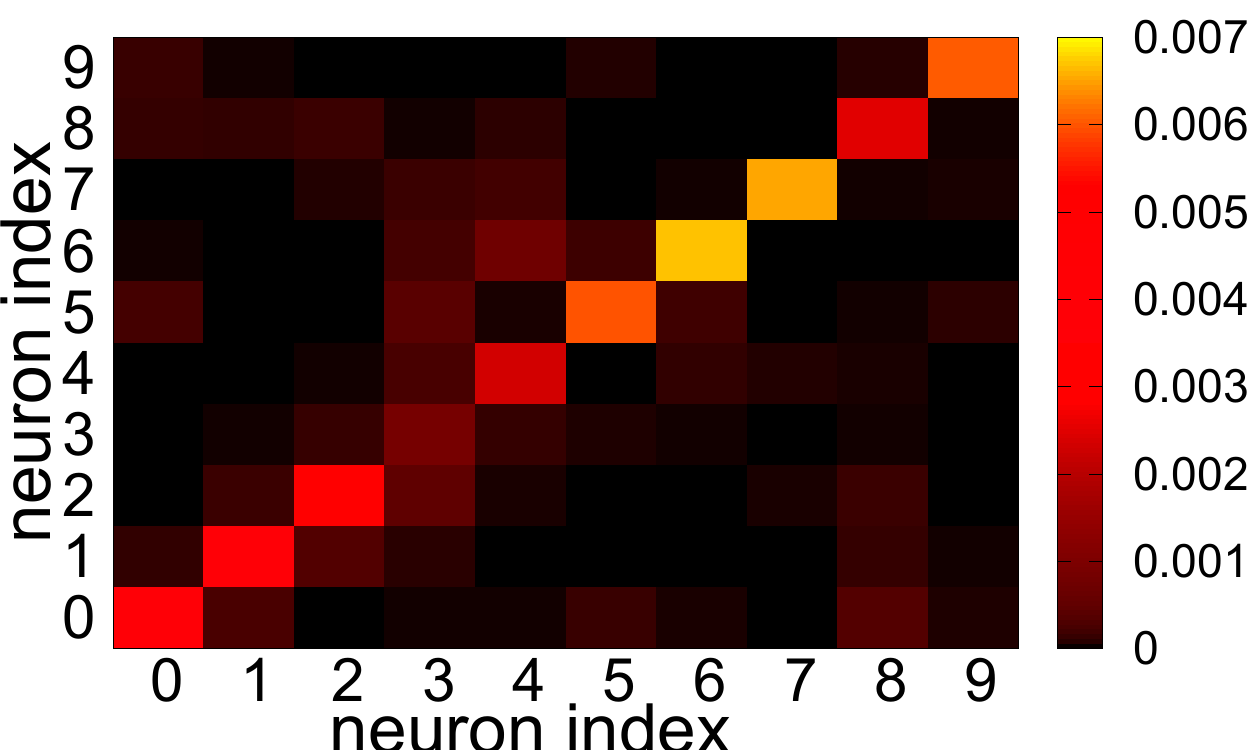}
\caption{\label{fig:corr}
Unnormalized covariance matrix between the activities of output neurons in a neural network trained on the multi-class classification task considered in this study. It can be observed that some neurons are not at all correlated, and thus are rarely active together, whereas other are correlated quite strongly. Our hypothesis is that this structure contains information about the task that can be used to further improve classification accuracy.
}
\end{figure}

\subsection{Contribution and novelty}
% what we do
% Novelty
This contribution proposes a pragmatic and application-oriented approach to MCC when performing an OVA decomposition approach. Instead of making a priori assumptions about distributions or conditional independence properties, it is attempted to learn such properties from data, and to exploit this knowledge for improved accuracy. For the concrete case of an MLP classifier trained on a difficult hand gesture recognition task, we investigate how the addition of another MLP stage that operates on the class output activities (COAs) of the first, can affect performance. Our basic assumption is that the selective inter-class correlations in the COAs, which can be observed in Fig.~\ref{fig:corr} and which we suppose will exist for any problem, contain useful information that the second MLP stage can extract and harness. Furthermore, we show for the well-known MNIST classification benchmark for handwritten digits, that the addition of this second stage does not always greatly improve performance, but that it causes no degradation either. 

The novelty of our approach lies in its simplicity and generality, as well as its practical applicability. As our classifier hierarchy
attempts to model the structure of the data by itself, no explicit assumptions need to be made by the user, other than issues of classifier design and parametrization for which standard techniques exist.

\section{Methods}

In this section, we mainly present two different training techniques, one of which can be extended $n$ times, depending on the problem. The dataset has to be prepared accordingly which is covered separately for each approach. Furthermore, we present the databases used for all experiments of Sec.~\ref{sec:experiments}
\begin{figure*}
\centering
\includegraphics[width=1.0\textwidth]{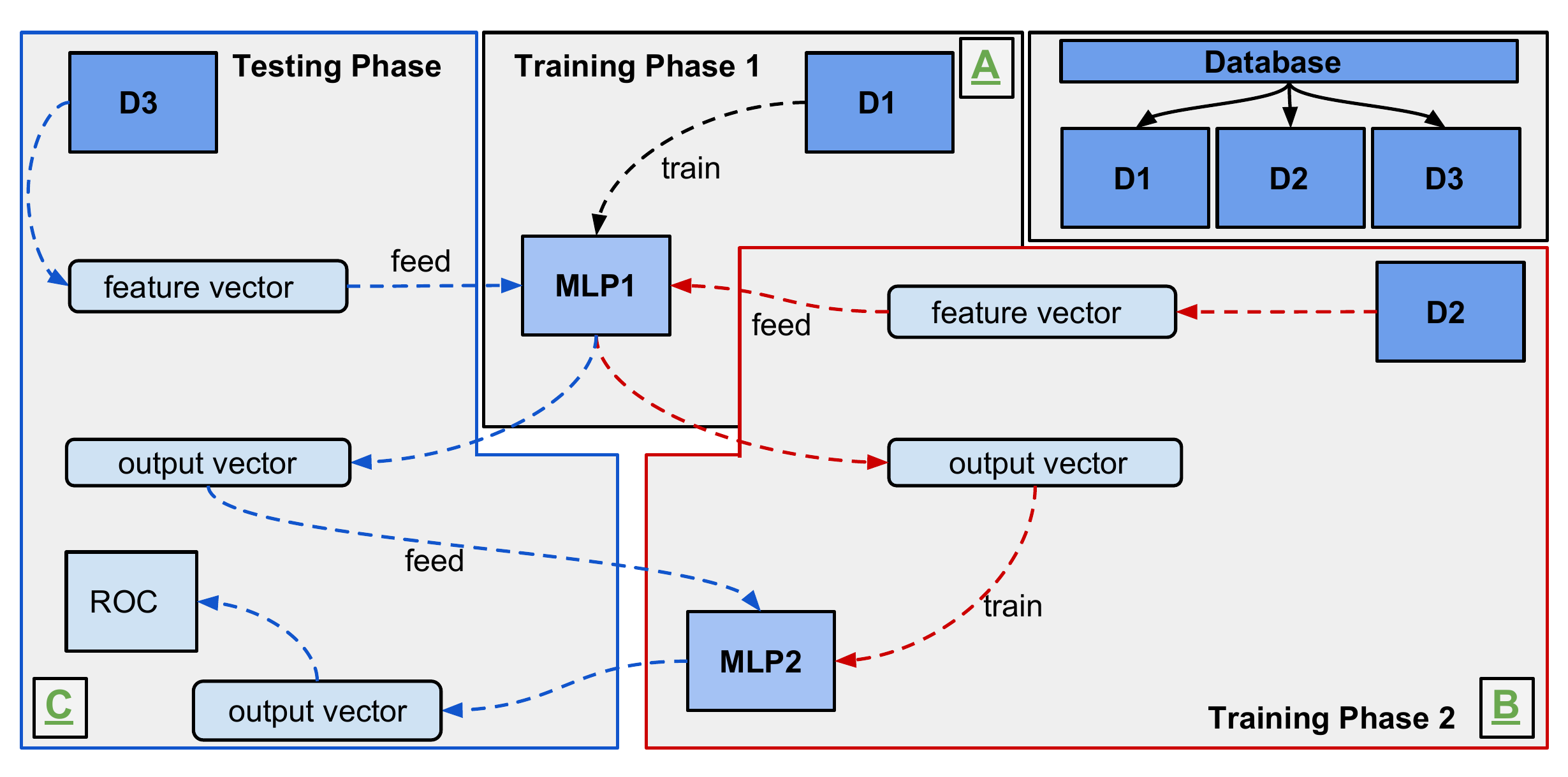}
\caption{\label{fig:procedure_output}
Training and testing procedure as described in Sec.~\ref{sec:outputs}. The whole database is randomly split into three subsets D1-D3. There are Training Phases 1 and 2 and one Test Phase (denoted A, B and C respectively) during which the MLPs are trained and evaluated. 
}
\end{figure*}
\begin{figure*}
\centering
\includegraphics[width=1.0\textwidth]{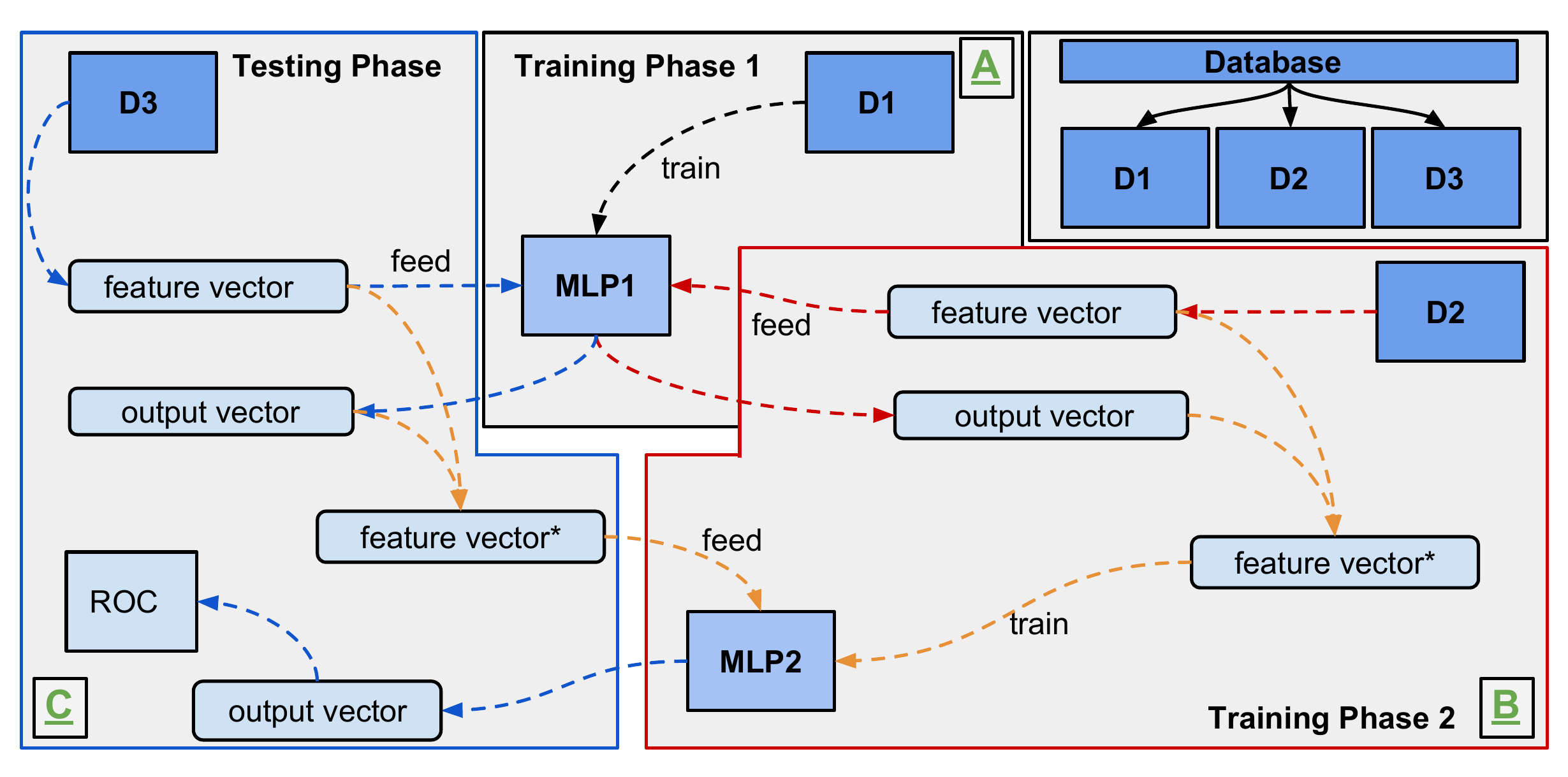}
\caption{\label{fig:procedure_inoutfusion}
Training and testing procedure as described in Sec.~\ref{sec:fusion}. The whole database is randomly split into three subsets D1-D3. There are Training Phases 1 and 2 and one Test Phase (denoted A, B and C respectively) during which the MLPs are trained and evaluated. The main differences are the fusion steps forming feature vector* from the original feature vector and the output vector. This occurs for MLP 2 in Training Phase B and Test Phase C, highlighted in orange (cf. with Fig.~\ref{fig:procedure_output}).
}
\label{fig:fusion}
\end{figure*}
\subsection{Training in output neurons}
\label{sec:outputs}
\begin{figure*}
\centering
\includegraphics[width=0.7\textwidth]{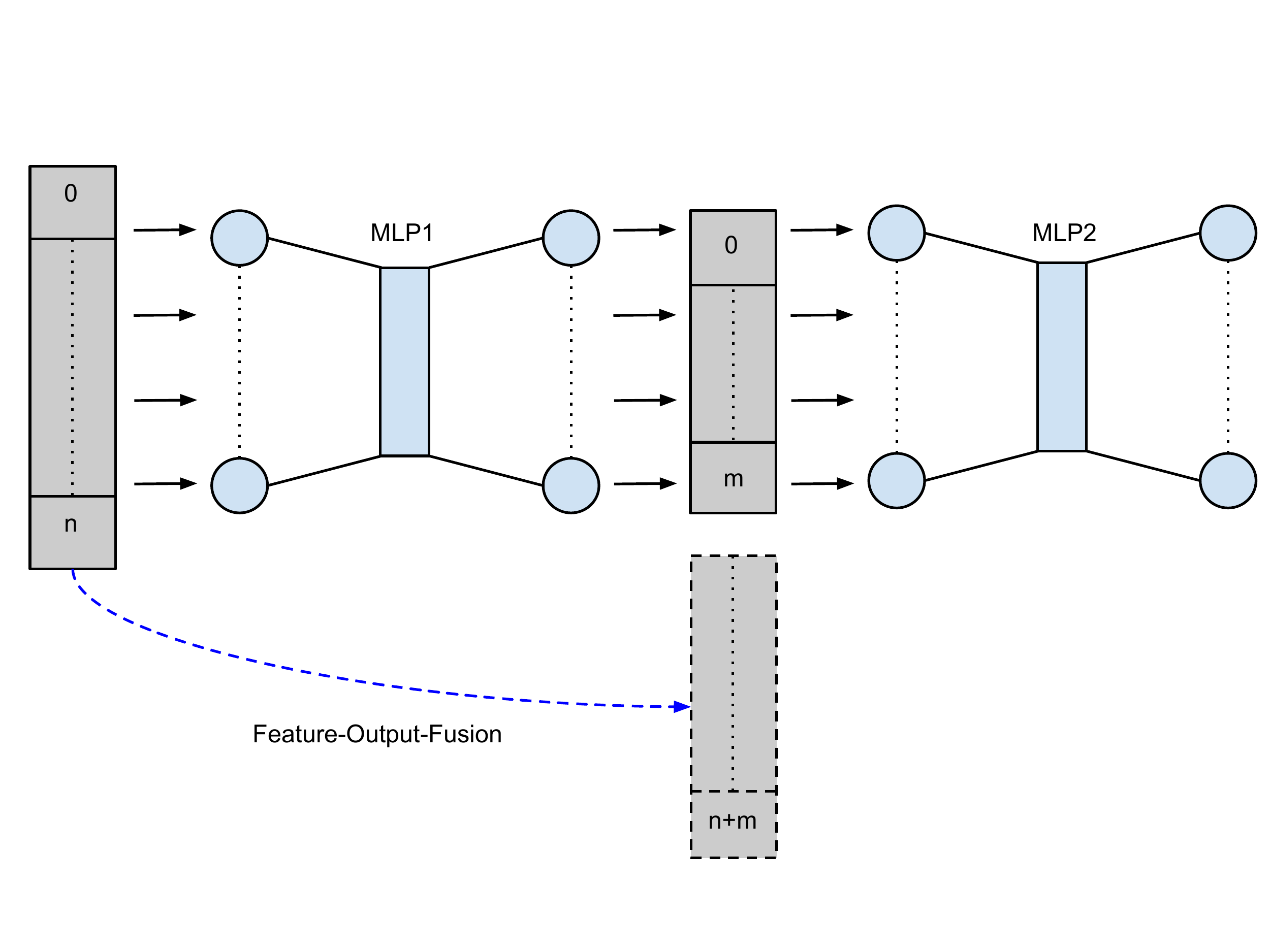}
\caption{\label{fig:mlpsetup}
Schematic procedure of sample propagation and fusion technique. MLP1 is always trained with the unchanged samples taken from the training data set. A sample, represented by a feature vector of length $n$, is fed into the MLP's input layer. After MLP1 has propagated the input and calculated each neuron's activation in the output layer, MLP2 is trained on the output vector of size $m$. Optionally, the output vector is fused with the feature vector itself (cf. Fig.~\ref{fig:procedure_inoutfusion}), forming  the new input of size $m+n$.}
\end{figure*}
This paragraph describes the cascading of two MLPs, where the basic idea is to let the first MLP classify a feature vector, and the second MLP the vector of output neurons of the first MLP. Training is performed sequentially, and care must be taken to prevent overfitting as each MLP is trained in a purely supervised fashion. 

The procedure consists of three stages A, B and C which are schematically depicted in Fig.~\ref{fig:procedure_output}. At first, the whole dataset has to be divided up randomly into three equally sized sets D1, D2 and D3. In an initial step, one MLP , here denoted MLP1, is trained with standard parameters (cf. Sec.~\ref{section:mlpstructure}) on D1. 
%We measure the difference of improvement of the mean squared error (MSE) during training of the MLP and end the training as soon as the improvement of the MSE falls below a certain threshold. 
Once MLP1 has converged, training of the second MLP, denoted MLP2, begins on dataset D2. The training is contrasted in such a way as now each individual training sample from D2 first has to be fed into MLP1. The input is propagated to the output layer and each output neuron makes a real-valued prediction for the possible class. These values form the output vector of length $m$ equalling the number of classes in the MCC task at hand. This output vector in turn corresponds to the input value of each neuron in the input layer of MLP2. Therefore, for this approach, the size of the input layer of MLP2 always equals the size of the MCC in the given task. This propagation of information is shown in Fig.~\ref{fig:mlpsetup}.
In this way, MLP2 is trained until convergence. The performance of our three-stage approach is then measured on dataset D3. Every sample is first fed into MLP1 which again calculates the values of its output neurons. These values are then, analogously to the training phase, presented as inputs to MLP2 which in turn calculates its own outputs. The determined class for a sample $S$ corresponds to the neuron with the highest activation in the output layer:
\begin{align}
class\{S\} &= \argmax{\{O_i\}} ,  0 \leq i \leq m,
\end{align}
$m$ being the number of classes of the MCC and $O_i$ representing the output neuron corresponding to class $i$. 
\subsection{Training in output neurons plus features}
\label{sec:fusion}
We extend the approach presented in the preceding section in such a way that the training of MLP2 (and also the evaluation) is performed on the neuron values of the output layer of MLP1 but also on the features of the sample itself, i.e., the input to MLP1.
The whole procedure again comprises three stages A, B and C and is schematically depicted in Fig.~\ref{fig:procedure_inoutfusion}, the main differences to the methodology shown in Fig.~\ref{fig:procedure_output} are highlighted.
The dataset is randomly split in an analogous manner to the approach described in Sec. \ref{sec:outputs} into three equally sized sets D1, D2 and D3. As before, MLP1 is trained on the whole dataset D1 until it converges. 

However the input for training MLP2 on D2 is now significantly different, comprising the feature vector of the sample plus the output vector of MLP1, 
 $\vec{v}_{out} = [O_0,...,O_n]$, n equalling the number of classes and $O_i$ being the activation of neuron $i$ in the output layer.
We thus form a new training input  $\bar{\vec{f}}$ resulting from the merging of the current feature vector $\vec{f}$ and the output vector $\vec{v}_{out}$ by simple concatenation. Its length len($\bar{\vec{f}}$) = len($\vec{f}$) + len($\vec{v}_{out}$) is determined by the size of the descriptor on the one hand and the number of classes on the other hand. 
%It therefore equals the size of the input layer of MLP2 as each element of $\bar{\vec{f}}$ forms the input of the corresponding neuron in the respective input layer of MLP2. The so-formed feature vector $\bar{\vec{f}}$ is now fed into the MLP2 for training. We proceed in this way to present each sample in D2 until a whole training epoch is presented and repeat the whole process until MLP2 has also converged. In order to test our fusion method on dataset D3 we proceed to first form the test sample from D3 by the fusion approach described in this chapter. The newly fused input  is the presented to MLP2 and the winner neuron corresponds to the neuron in the output layer with the highest activation.\\
Fig.~\ref{fig:mlpsetup} shows the propagation of the features and the formation of the input, in this case the Feature-Output-Fusion. Opposed to the procedure described in Sec.~\ref{sec:outputs} the input now has length $n+m$, $n$ being the size of the feature vector and $m$ the size of the MCC.

A variation of this approach with only a slight modification of the original algorithm is simply achieved by training two separate MLPs (MLP1-A, MLP1-B) on D1. Now we are able to concatenate both MLP outputs with the feature vector coming from training and test set D2 and D3 respectively. The main difference is that each sample, in training and testing phase, is now presented to both MLP1-A and MLP1-B which propagate their inputs so that both outputs generated from each MLP can be concatenated with the feature vector. The length of the newly formed feature vector now of course differs, depending on the size of the classification task, i.e. for the approach just described it is formed as  len($\bar{\vec{f}}$) = len($\vec{f}$) + $k*$len($\vec{v}_{out}$), with $k$ being the number of different MLPs trained on D1 or on separate datasets. 
\subsection{Hand Gesture Database and Descriptors}
\begin{figure*}
	
	\includegraphics[width=0.7\textwidth]{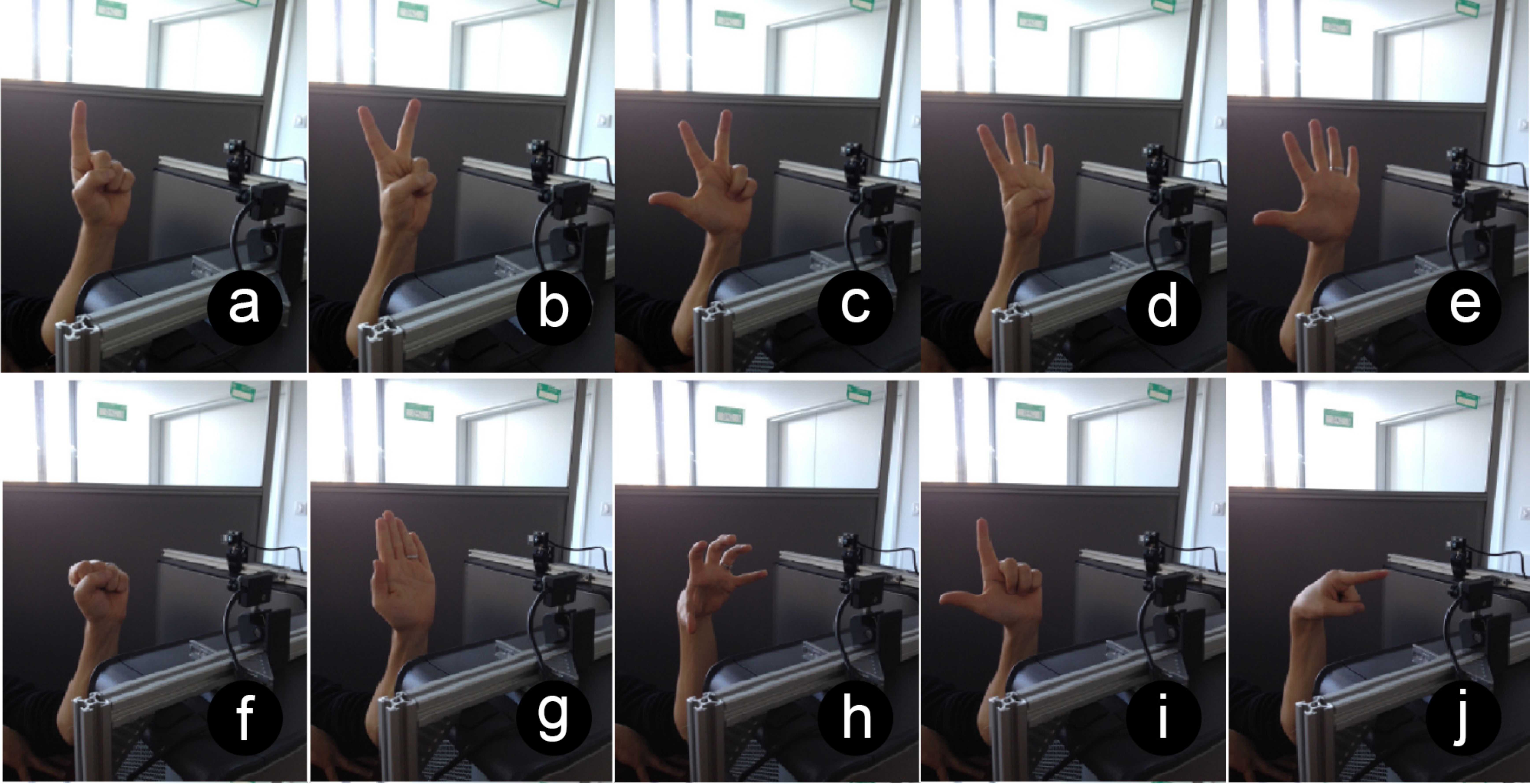}
	\centering
	\caption{ The hand gesture database consisting of 10 different static hand poses.}

	\label{figure:db}
\end{figure*}
We record data from 16 persons, each displaying 10 different hand poses (cf. Figure \ref{figure:db}). One data sample is stored as a so-called point cloud describing the hand shape of a person by a vector of real-valued x-y-z coordinates. For each gesture, 3000 samples are recorded, summing up to 30000 samples per person and a total database of 480000 samples. In order to induce some variance into the data, during the recording phase each participant is asked to rotate and translate their hand in all possible directions. Moreover, to tackle the task of scaling, for each gesture we define 3 different distance ranges, in which the participant is asked to perform the hand gesture in order to ensure sufficient sample coverage for various distances. To sum up, this results in an alphabet of ten hand poses: Counting from 1-5 and \textit{fist}, \textit{stop}, \textit{grip}, \textit{L}, \textit{point} denoted by \textit{a-j} (cf. Figure \ref{figure:db}).\\
Each sample originates from raw point cloud data which is transformed via a global Point Cloud Descriptor into a histogram of fixed size. The descriptor depicts the global shape of the cloud via the relationship of the angles calculated from a sample subset of point pairs as well as the distance between a point pair. The histogram calculated in this manner forms the feature vector i.e. the input for the MLPs in the early training and late fusion stages described in Sec.~\ref{sec:outputs} and Sec.~\ref{sec:fusion}. For a more in-depth specification of this feature transformation please refer to \cite{kopinski2014neural}.
\section{MLP structure}
\label{section:mlpstructure}
Within the frame of the experiments, each MLP comprises one input, hidden and output layer, each layer being fully connected to its successor (cf. Fig.~\ref{fig:mlpsetup}). The output layer depends on the size of the classification task as each class is represented by a neuron. Depending on the training technique used, see Secs.~\ref{sec:outputs}, \ref{sec:fusion}, the input layer varies in size. During the first stage, the input layer of the first MLP always equals the size of the feature vector. For the second stage, this varies depending on the technique employed as the input here is either formed by the output vector alone or the output vector concatenated with the feature vector. Therefore the input layer is of size $n$ for MLP1 during training and testing and of size $m$ or $m+n$ for MLP2, the number of classes or the number of classes + length of feature vector respectively, depending on the technique. When employing the extension of the method described in Sec.~\ref{sec:fusion}, the size of the input layer increases to $n+k*m$, $k$ being the number of individual MLPs trained on a dataset.

For the experiments, varying hidden layer sizes were tested within range of $[40,150]$ neurons, having a noteworthy but not excessive effect on the results. Depending on the size of the input layer, which can vary due to the techniques described in this paper, choosing a different number of hidden neurons may be beneficial. Since this is also beyond the scope of this contribution we only state that a  proper parameter search may lead to some improvement in terms of classification performance.
\section{Experiments}
\label{sec:experiments}
All methods were implemented using the FANN library (see \cite{nissen2003implementation}). The training algorithm is RPROP, the activation function is the sigmoid function in both hidden and output layers. The rest of the MLP parameters are standard parameters we don't vary during the course of our experiments, as we conducted a series of initial test runs to determine proper parameters for the described methodology. A complete parameter search is beyond the scope of this contribution.

We have conducted a series of experiments, all divided into two different testing phases (Phase 1 and Phase 2) in order to be able to directly compare the effects of our techniques on the classification performance of each MLP. The results are depicted in the confusion matrices which allow us to compare the overall performance of the MLPs, the performance on each individual class as well as the correlations between all classes. 
\subsection{Experiment 1 - output neurons only}
\label{sec:exp_1}

\begin{figure}
\centering
\includegraphics[width=0.5\textwidth]{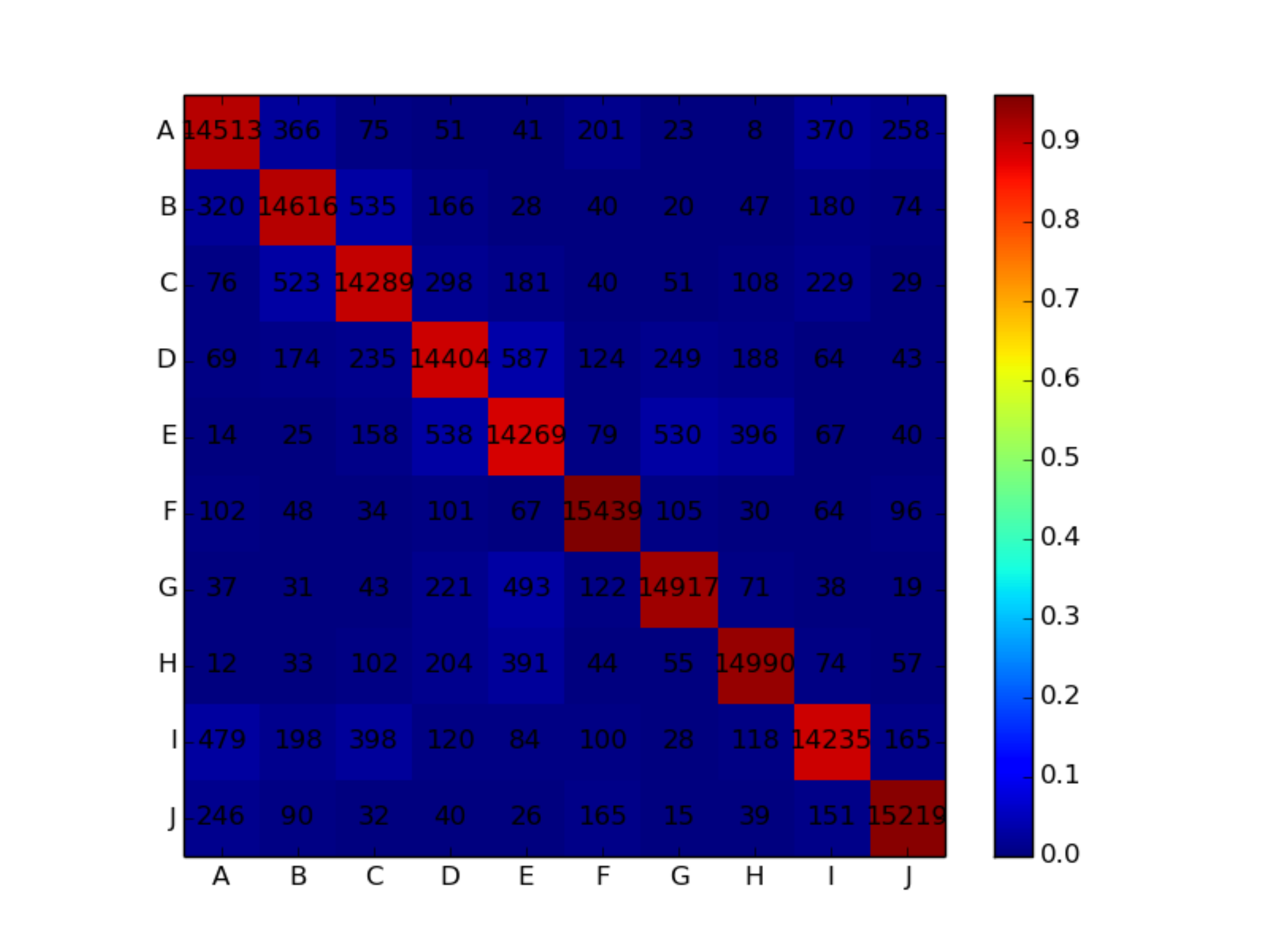}
\caption{\label{fig:non1}
Experiment 1, Phase 1: Confusion Matrix for the first MLP trained only on the feature vectors.
}
\end{figure}

\begin{figure}
\centering
\includegraphics[width=0.5\textwidth]{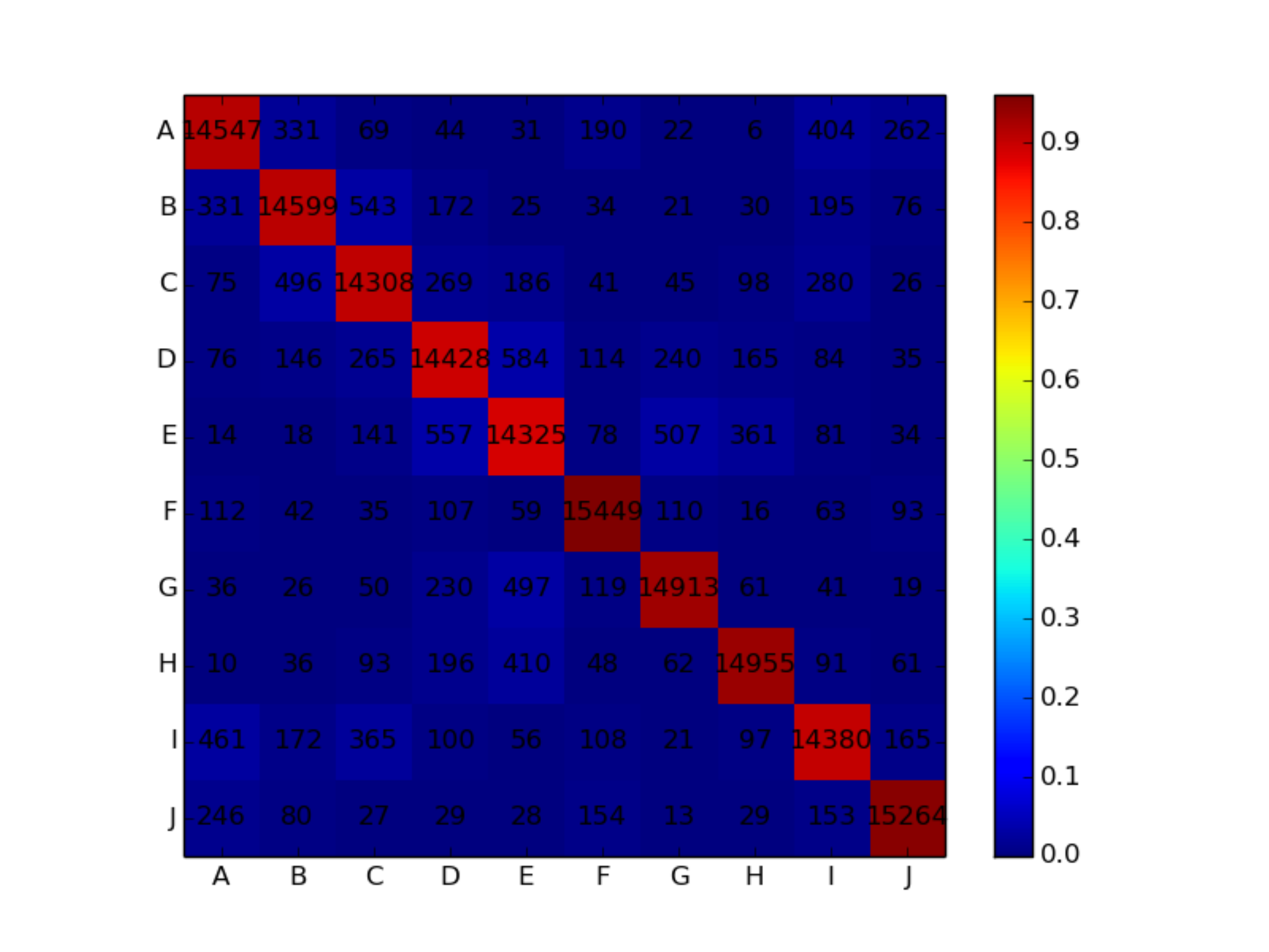}
\caption{\label{fig:non2}
Experiment 1, Phase 2: Confusion Matrix for the second MLP trained only on the net outputs of the first MLP.
}
\end{figure}

In the first experiment we test the effect of the technique described in Sec.~\ref{sec:outputs}. We randomly split the whole database into three subsets D1-D3 (two for training and one for testing). Both MLPs have 100 neurons in the hidden layer and were trained on subset D1 and D2 respectively until they converged. The results can be seen in Fig.~\ref{fig:non1} and Fig.~\ref{fig:non2}. The overall performance of MLP1 (trained on the feature vectors) is at around 91.80\% and at around 91.98\% for MLP2  (trained on the output vector of MLP2) which is an improvement of around 0.2\%. Overall, moderate improvements can be observed in nearly all cases, two cases are subject to negligible decrease in performance ($<$ 0,001\%).

\subsection{Experiment 2 - output neurons plus features}
\label{sec:exp_2}

\begin{figure}
\centering
\includegraphics[width=0.5\textwidth]{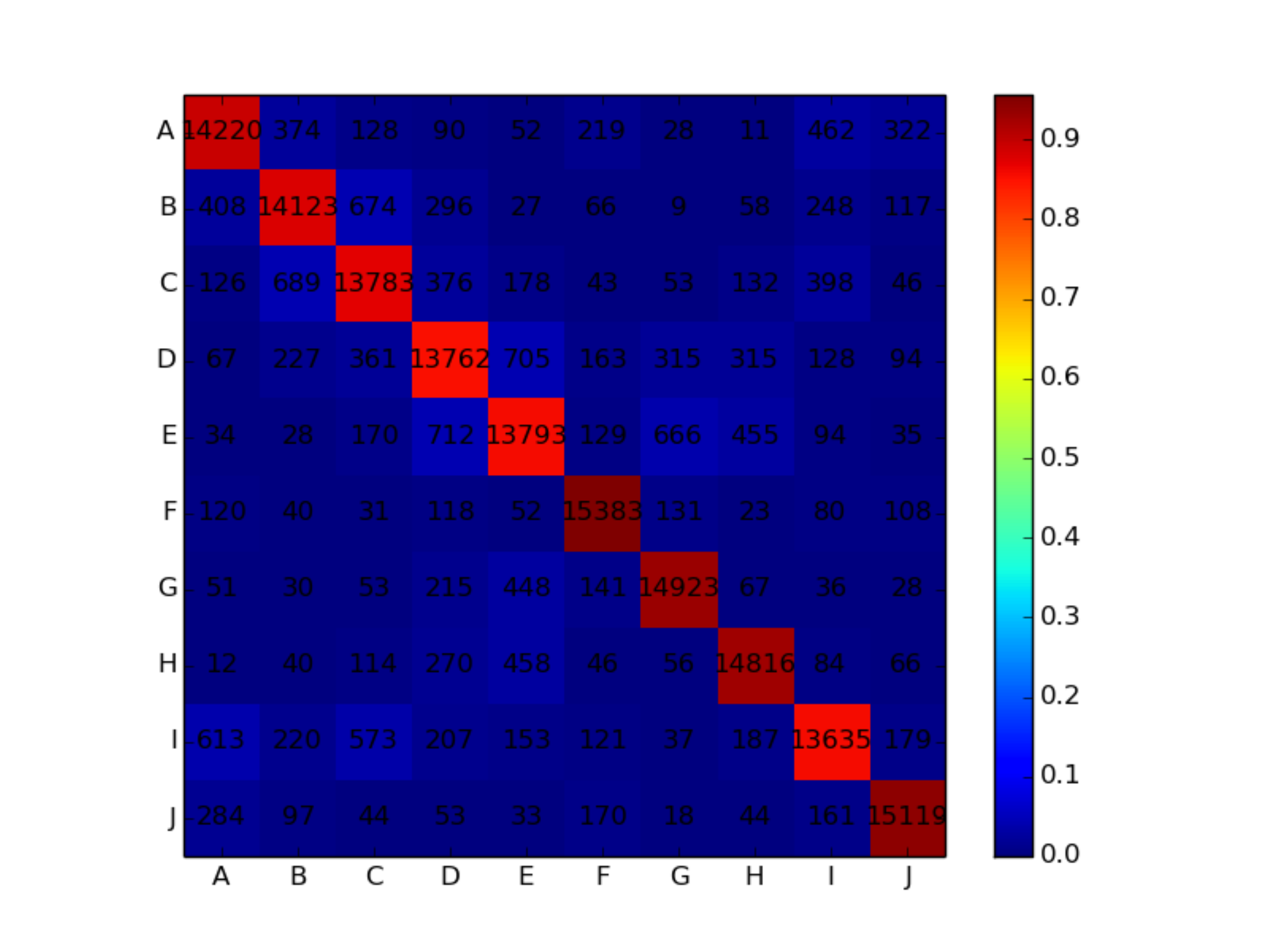}
\caption{\label{fig:fusion1_1}
Experiment 2, Phase 1: Confusion Matrix for the first MLP trained only on the feature vectors.
}
\end{figure}

\begin{figure}
\centering
\includegraphics[width=0.5\textwidth]{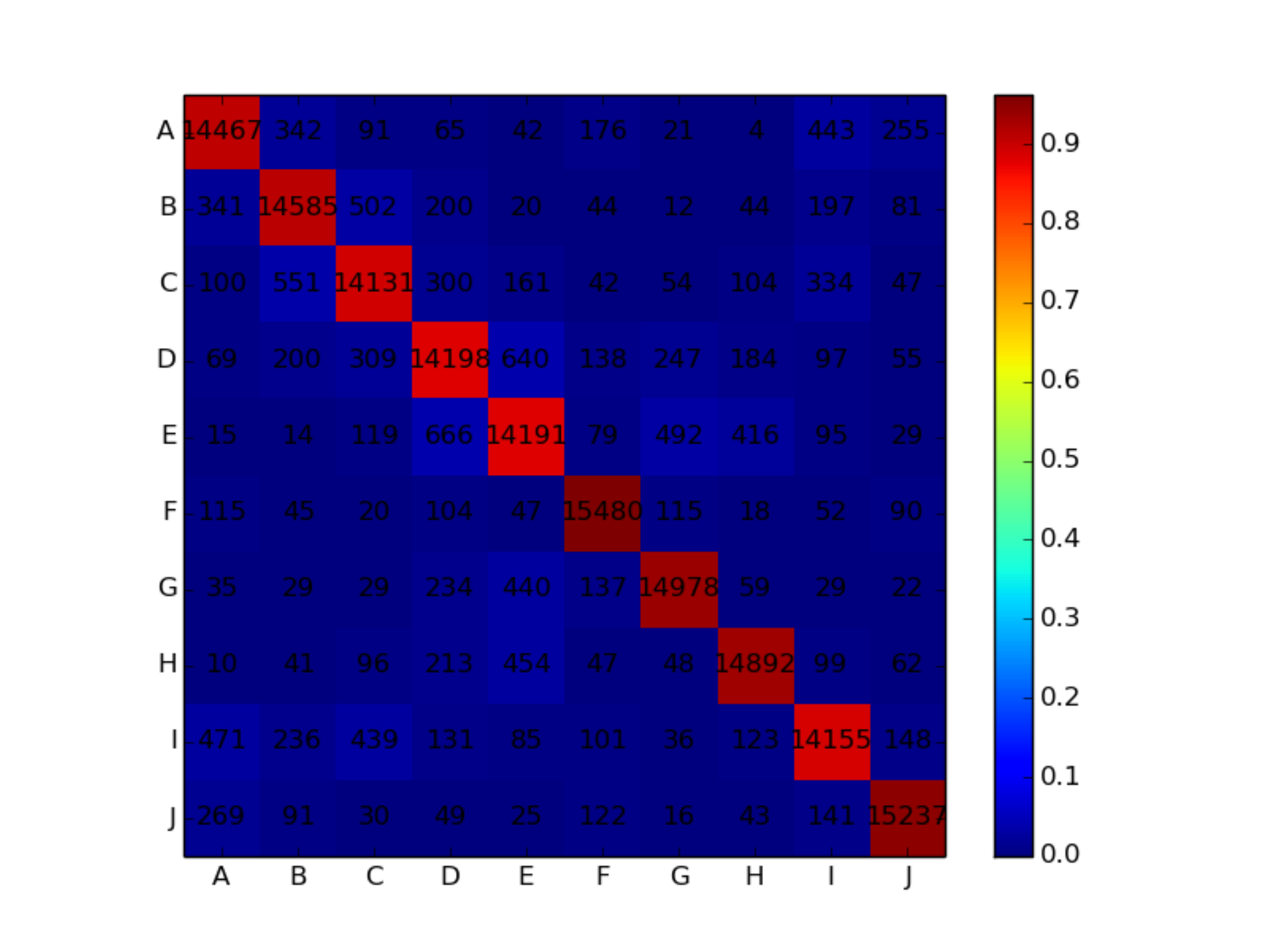}
\caption{\label{fig:fusion1_2}
Experiment 2, Phase 2: Confusion Matrix for the second MLP trained on the fused feature vectors (features + outputs).
}
\end{figure}

In the second experiment we test the effect of the technique described in Sec.~\ref{sec:fusion}. We randomly split the whole database into three subsets D1-D3 (two for training and one for testing). Both MLPs have 80 neurons in the hidden layer and were trained on subset D1 and D2 respectively until they converged. The results can be seen in Fig.~\ref{fig:fusion1_1} and Fig.~\ref{fig:fusion1_2}. The overall performance of MLP1 (trained on the feature vectors) is at around 90.0\% and at around 91.0\% for MLP2  (trained on the fused vector) which is an improvement of around 1.0\%.

\begin{table}[h]
\resizebox{\columnwidth}{!}{
\begin{tabular}{|c|c|c|c|c|c|c|c|c|c|c|}
\hline
     & A    & B    & C    & D    & E    & F    & G    & H    & I    & J    \\ \hline
MLP1 & 89\% & 88\% & 87\% & 85\% & 86\% & 96\% & 93\% & 93\% & 86\% & 94\% \\ \hline
MLP2 & 91\% & 91\% & 89\% & 88\% & 88\% & 96\% & 94\% & 93\% & 89\% & 95\% \\ \hline
\end{tabular}
}
\caption{Classification results for MLP1 and MLP2. The ten classes are named A-J.}
\label{tab:exp1}
\end{table}

Tab.~\ref{tab:exp1} gives more insight into the improvements of classification performance related to each individual gesture class. There is a performance increase for all cases (which is below 0.5\% in cases F and H) and ranges between 1-3\% for all other remaining cases. Most notable the presented approach significantly boosts performance in situations where MLP1 performs poorly (cf. cases D + I) as opposed to little improvement in cases where MLP1 already performs well (e.g. cases F + J).\\ 
When comparing the confusion matrices of these cases one can see that the improvement stems mainly from those classes which contained most false positives, i.e. class I was most likely to be mistaken for class A or C (cf. Fig.~\ref{fig:fusion1_1} and Fig.~\ref{fig:fusion1_2}). The number of false positives for this specific example drops by a rate of $>$20\% which is significant as it allows an improved disambiguation procedure ($613 \rightarrow 471$ and $573 \rightarrow 439$ respectively).

\subsection{Experiment 3 - output neurons plus features with multiple MLPs}
\label{sec:exp_3}

\begin{figure}
\centering
\includegraphics[width=0.5\textwidth]{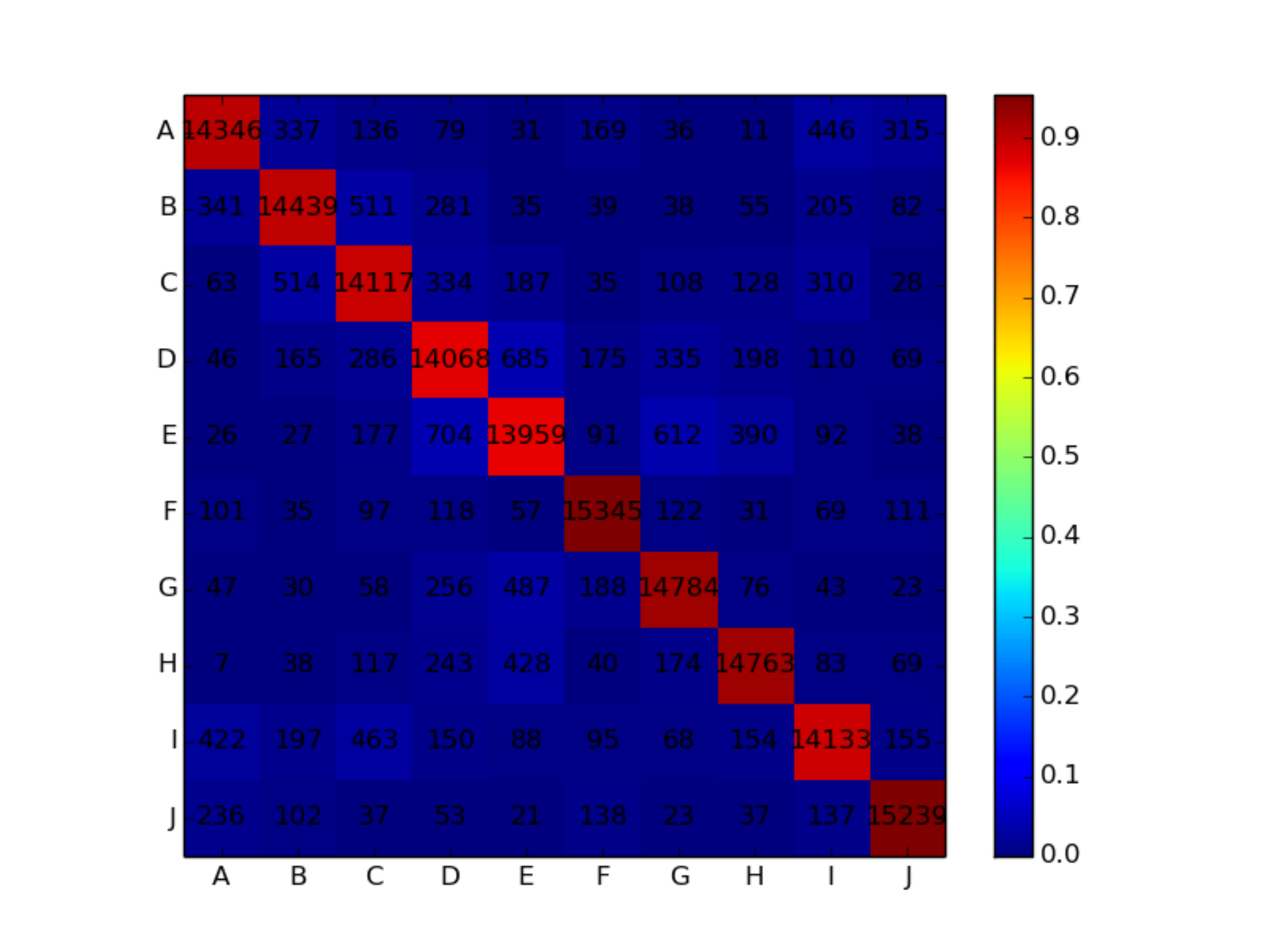}
\caption{\label{fig:fusion2_1}
Experiment 3, Phase 1: Confusion Matrix for the first MLP trained only on the feature vectors.
}
\end{figure}

\begin{figure}
\centering
\includegraphics[width=0.5\textwidth]{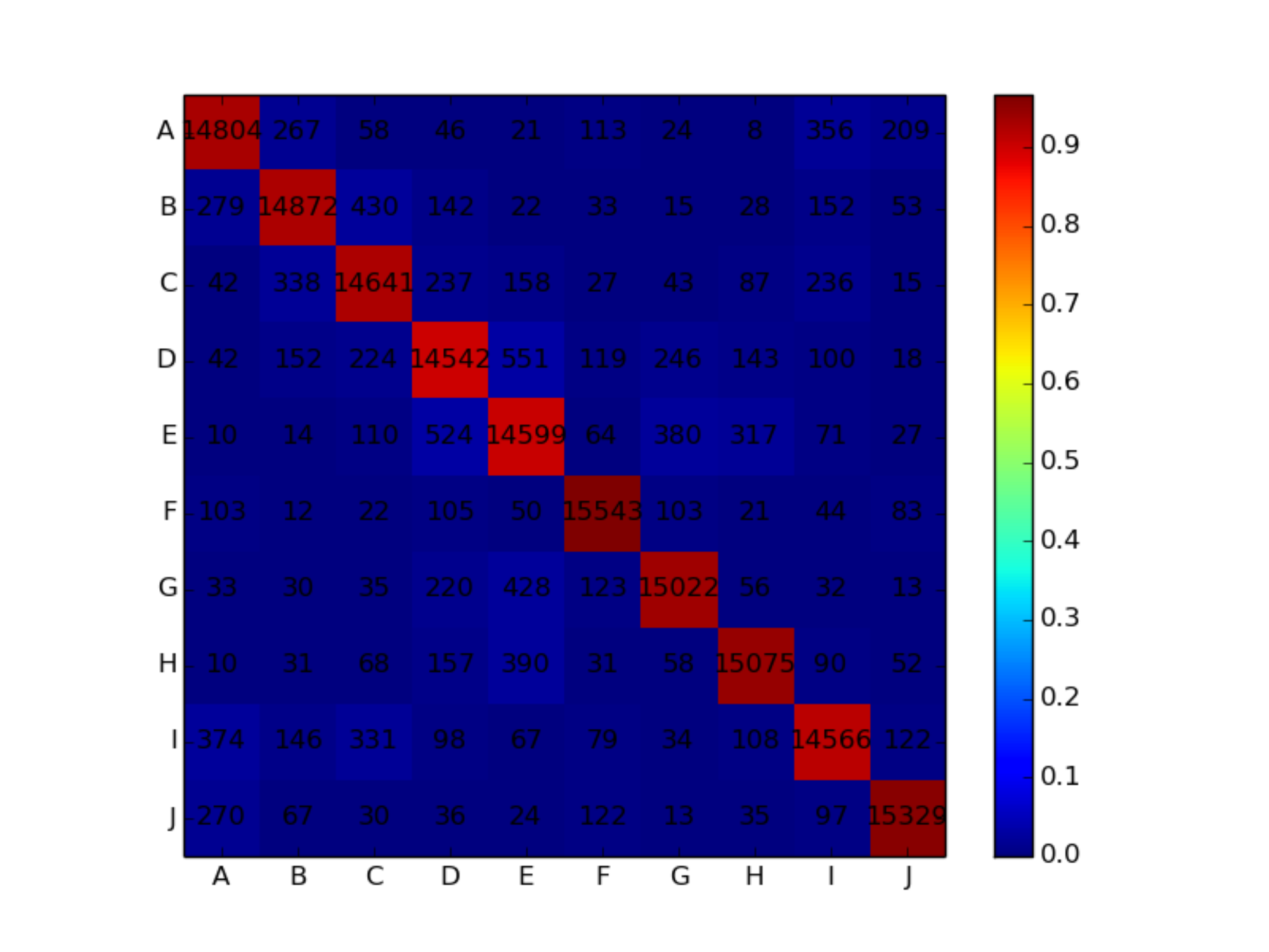}
\caption{\label{fig:fusion2_2}
Experiment 3, Phase 2: Confusion Matrix for the second MLP from the extended fusion technique, i.e. the fused feature vector coming the original feature vector concatenated with the outputs from two separately trained MLPs.
}
\end{figure}

The third experiment evaluates the effect of the extended technique described in Sec.~\ref{sec:fusion}. We randomly split the whole database into three subsets D1-D3 (two for training and one for testing). The main difference here is resembled by the fact that we used two MLPs (MLP1-A, MLP1-B) trained on set D1, instead of just one. Therefore we first have to feed every sample into both MLPs and calculate their output vectors. These are in turn then concatenated with the feature vector to form the new input vector for MLP2 during training and testing. The results of this extended technique are shown in Fig.~\ref{fig:fusion2_1} and Fig.~\ref{fig:fusion2_2}. The overall performance of MLP1 (trained on the feature vectors) is at around 91.0\% and at around 93.0\% for MLP2 (trained on the fused vector) which is an overall improvement of around 2.0\%.

\begin{table}[h]
\resizebox{\columnwidth}{!}{
\begin{tabular}{|l|l|l|l|l|l|l|l|l|l|l|}
\hline
     & A    & B      & C      & D      & E      & F      & G      & H      & I      & J      \\ \hline
MLP1 & 90\% & 90\%   & 89\%   & 87\%   & 87\%   & 95\%   & 92\%   & 92\%   & 89\%   & 95\%   \\ \hline
MLP2 & 93\% & 93\%   & 93\%   & 90\%   & 91\%   & 97\%   & 94\%   & 94\%   & 91\%   & 96\% \\ \hline
\end{tabular}
}
\caption{Classification results for MLP1 and MLP2. The ten classes are named A-J.}
\label{tab:exp2}
\end{table}

Individual improvements range from 2-4\% for all classes (cf. Tab.\ref{tab:exp2}). As before (cf. Sec.~\ref{sec:exp_3}) misclassification rates drop around 20\% - 25\% for the most difficult cases (compare class E: $704 \rightarrow 524$ and $612 \rightarrow 380$ for cases D and G respectively).

\subsection{Experiment 4 - MNIST}
\label{sec:exp_4}
As a supplementary exercise, we test our approach on the well-known MNIST hand-written digit database\cite{lecun-01a}, using the fusion strategy from Sec.~\ref{sec:fusion}. We observe, without any parameter tuning, a classification around 93\% which does not noticeably improve by adding the second MLP.
On the other hand, no degradation of performance is observed either. What seems to be the problem here is that, being forced to divide the dataset into three parts, we can use less training examples than other approaches can, which maybe explains the lack of improvement. 
\section{Discussion and Outlook}
In this article, we present a multi-class classification scheme that is intended to be useful in practical applications. As it does not make explicit assumptions about the nature of classification tasks, it contains no additional parameters 
beyond those that would have to be tuned any case for binary classifier training, in this case MLPs. No significant theoretical modeling of the classification task at hand needs to be performed as we rely on the capacity of the second MLP to extract those properties to the best of its capabilities. Furthermore, when observing the result, we find that the {\it overall} improvements are modest, i.e., in the range of $\le 5\%$. However, in an application it is often not the overall
classification rate that is of importance, but the worst case, that is to say, the behavior of the classifier for specific "difficult" classes. Here, we observe 
a strong benefit from using our two-stage approach as the performance on some classes improves by $>20\%$ which is highly significant in practice. Further strengthening the link to practical applicability, we observe that the additional computational cost of adding the second MLP is virtually non-existent, or rather, very hard to measure due to the efficient C implementation. We therefore obtain significant gains in applicability at negligible computational cost, which is always an important point in practice, especially in a vehicular context where we apply this technique for the purposes of human-machine interaction. 

The reason for the improvement as far as "difficult" classes are concerned, probably stems from the fact that those classes are very similar to certain others and thus are often confused by MLP1. The second MLP can presumably detect such confusion events by specific patterns of activated output neurons in MLP1, and correct the decision. The fact of providing the original feature vector seems to have a beneficial effect on this technique. 

A critical point of the presented approach is its hunger for data: as we split the original dataset into three parts, instead of two as is usually the case, 
we have less data to train our classifiers with, potentially incurring a loss of performance. In order to remedy this, we are currently studying 
the question of how to perform the presented scheme with only two data sets, one for training and one for testing. This would involve training MLP1 and MLP2, in a supervised fashion, on the same dataset, which we consider problematic for reasons of overfitting. Nevertheless, initial test have shown that ge\-ne\-ra\-li\-za\-tion performance is not in the least affected by this, so we will pursue this avenue of research further, possible with the aid of advanced regularization methods. After all, deep belief networks train their layers one after the other, each layer on the outputs of the previous one, all on the same training set. 

A further critical point is the lack of generality of the presented experiments: in future work, we will definitely apply this method to a multitude of other datasets in order to better validate its worthwhileness. This will also allow us to determine whether the approach gives higher performance gains for rather simple problems (as treated here), or for very hard problems. 
%\section*{Acknowledgment}


\bibliographystyle{IEEEtran}
\bibliography{allPapers.bib}

\begin{thebibliography}{10}
\providecommand{\url}[1]{#1}
\csname url@rmstyle\endcsname
\providecommand{\newblock}{\relax}
\providecommand{\bibinfo}[2]{#2}
\providecommand\BIBentrySTDinterwordspacing{\spaceskip=0pt\relax}
\providecommand\BIBentryALTinterwordstretchfactor{4}
\providecommand\BIBentryALTinterwordspacing{\spaceskip=\fontdimen2\font plus
\BIBentryALTinterwordstretchfactor\fontdimen3\font minus
  \fontdimen4\font\relax}
\providecommand\BIBforeignlanguage[2]{{%
\expandafter\ifx\csname l@#1\endcsname\relax
\typeout{** WARNING: IEEEtran.bst: No hyphenation pattern has been}%
\typeout{** loaded for the language `#1'. Using the pattern for}%
\typeout{** the default language instead.}%
\else
\language=\csname l@#1\endcsname
\fi
#2}}

\bibitem{Bishop2006}
C.~Bishop, \emph{Pattern recognition and machine learning}.\hskip 1em plus
  0.5em minus 0.4em\relax Springer-Verlag, New York, 2006.

\bibitem{guermeur}
Y.~Guermeur, ``Svm multiclasses, th\'eorie et applications (in french),''
  Universit\'e Nancy 1, Tech. Rep. {HDR} thesis, 2007.

\bibitem{msvm-1}
K.~Crammer and Y.~Singer, ``On the learnability and design of output codes for
  multiclass problems,'' \emph{Machine Learning}, vol.~47, no.~2, pp. 201--233,
  2002.

\bibitem{msvm-2}
Y.~Lee, Y.~Lin, and G.~Wahba, ``Multicategory support vector machines : Theory
  and application to the classification of microarray data and satellite
  radiance data,'' \emph{Journal of the American Statistical Association},
  vol.~99, no. 465, pp. 67--81, 2004.

\bibitem{msvm-3}
I.~Tsochantaridis, T.~Hofmann, T.~Joachims, and Y.~Altun, ``Support vector
  machine learning for interdependent and structured output spaces,'' in
  \emph{{ICML}'04}, 2004, pp. 823--830.

\bibitem{msvm-4}
D.~Anguita, S.~Ridella, and D.~Sterpi, ``A new method for multiclass support
  vector machines,'' in \emph{{IJCNN}'04}, 2004, pp. 407--412.

\bibitem{zhu2005kernel}
J.~Zhu and T.~Hastie, ``Kernel logistic regression and the import vector
  machine,'' \emph{Journal of Computational and Graphical Statistics}, vol.~14,
  no.~1, 2005.

\bibitem{ova-1}
B.~Schölkopf, C.~Burges, and V.~Vapnik, ``Extracting support data for a given
  task,'' 1995, pp. 252--257.

\bibitem{ova-2}
R.~Rifkin and A.~Klautau, ``In defense of one-vs-all classification,''
  \emph{Journal of Machine Learning Research}, vol.~5, pp. 101--141, 2003.

\bibitem{ovo-1}
J.~Friedman, ``Another approach to polychotomous classification,'' Department
  of Statistics, Stanford University, Tech. Rep., 1996.

\bibitem{ovo-2}
E.~Mayoraz and E.~Alpaydin, ``Support vector machines for multi-class
  classification,'' {IDIAP}, Tech. Rep. 98-06, 1998.

\bibitem{ovo-3}
U.~Kreÿel, ``Pairwise classification and support vector machines,'' in
  \emph{Advances in Kernel Methods, Support Vector Learning}.\hskip 1em plus
  0.5em minus 0.4em\relax The MIT Press, Cambridge, MA, 1999, pp. 255--268.

\bibitem{ovo-4}
T.~Hastie and R.~Tibshirani, ``Classification by pairwise coupling,'' \emph{The
  Annals of Statistics}, vol.~26, no.~2, pp. 451--471, 1998.

\bibitem{ovo-5}
M.~Moreira and E.~Mayoraz, ``Improved pairwise coupling classification with
  correcting classifiers,'' in \emph{{ECML}'98}, 1998, pp. 160--171.

\bibitem{ovo-6}
Z.~Li, S.~Tang, and S.~Yan, ``Multi-class svm classifier based on pairwise
  coupling,'' in \emph{SVM'02}, 2002, pp. 321--333.

\bibitem{ovo-7}
J.~F\"urnkranz, ``Round robin classification,'' \emph{Journal of Machine
  Learning Research}, vol.~2, pp. 721--747, 2011.

\bibitem{ovo-8}
C.~Angulo, X.~Parra, and A.~Català, ``K-{SVCR}. a support vector machine for
  multi-class classification,'' \emph{Neurocomputing}, vol.~55, no. 1-2, pp.
  57--77, 2003.

\bibitem{trees-1}
K.~B. Deslem and Y.~Bennani, ``Dendogram-based svm for multi-class
  classification,'' \emph{Journal of Computing and Information Technology -
  {CIT}}, vol.~14, no.~4, pp. 283--289, 2006.

\bibitem{trees-2}
J.~Platt, N.~Cristianini, , and J.~Shawe-Taylor, ``Large margin dags for
  multiclass classification,'' in \emph{{NIPS}}, 2000, pp. (547--553.

\bibitem{platt1999probabilistic}
J.~Platt \emph{et~al.}, ``Probabilistic outputs for support vector machines and
  comparisons to regularized likelihood methods,'' \emph{Advances in large
  margin classifiers}, vol.~10, no.~3, pp. 61--74, 1999.

\bibitem{kopinski2014neural}
T.~Kopinski, S.~Geisler, U.~Handmann, and A.~Gepperth, ``Neural network based
  data fusion for hand pose recognition with multiple tof sensors,'' in
  \emph{International Conference on Artificial Neural Networks 2014}, 2014.

\bibitem{nissen2003implementation}
S.~Nissen, ``Implementation of a fast artificial neural network library
  (fann),'' \emph{Report, Department of Computer Science University of
  Copenhagen (DIKU)}, vol.~31, 2003.

\bibitem{lecun-01a}
Y.~LeCun, L.~Bottou, Y.~Bengio, and P.~Haffner, ``Gradient-based learning
  applied to document recognition,'' in \emph{Intelligent Signal
  Processing}.\hskip 1em plus 0.5em minus 0.4em\relax IEEE Press, 2001, pp.
  306--351.

\end{thebibliography}

\end{document}